\documentclass[11pt,a4paper]{article}
\usepackage[hyperref]{acl2018}
\usepackage{times}
\usepackage{latexsym}
\usepackage{amssymb}
\usepackage{url}
\usepackage{graphicx}
\usepackage{amsmath}
\usepackage{breqn}
\usepackage{algorithm}  
\usepackage{algorithmicx}  
\usepackage{algpseudocode}

\aclfinalcopy 
\def\aclpaperid{601} 

\title{Robust Distant Supervision Relation Extraction via Deep Reinforcement Learning}

\author{Pengda Qin$^\sharp$, Weiran Xu$^\sharp$, William Yang Wang$^\flat$ \\
  $^\sharp$Beijing University of Posts and Telecommunications, China \\
  $^\flat$University of California, Santa Barbara, USA\\
  {\tt \{qinpengda, xuweiran\}@bupt.edu.cn} \\
  {\tt \{william\}@cs.ucsb.edu} \\
  }

\date{}

\hypersetup{draft}

\begin{document}
\maketitle
\begin{abstract}
Distant supervision has become the standard method for relation extraction. However, even though it is an efficient method, it does not come at no cost---The resulted distantly-supervised training samples are often very noisy. To combat the noise, most of the recent state-of-the-art approaches focus on selecting one-best sentence or calculating soft attention weights over the set of the sentences of one specific entity pair. However, these methods are suboptimal, and the {\bfseries false positive} problem is still a key stumbling bottleneck for the performance. We argue that those incorrectly-labeled candidate sentences must be treated with a hard decision, rather than being dealt with soft attention weights. To do this, our paper describes a radical solution---We explore a deep reinforcement learning strategy to generate the false-positive indicator, where we automatically recognize false positives for each relation type without any supervised information. Unlike the removal operation in the previous studies, we redistribute them into the negative examples. The experimental results show that the proposed strategy significantly improves the performance of distant supervision comparing to state-of-the-art systems.
\end{abstract}

\section{Introduction}
Relation extraction is a core task in information extraction and natural language understanding. The goal of relation extraction is to predict relations for entities in a sentence~\cite{zelenko2003kernel,bunescu2005subsequence,guodong2005exploring}. 
For example, given a sentence \emph{``\textbf{Barack Obama} is married to \textbf{Michelle Obama}.''}, a relation classifier aims at predicting the relation of \emph{``\textbf{spouse}''}. 
In downstream applications, relation extraction is the key module for constructing knowledge graphs, and it is a vital component of many natural language processing applications such as structured search, sentiment analysis, question answering, and summarization. 
\begin{figure}[t]
\begin{center}
\includegraphics[width=8cm]{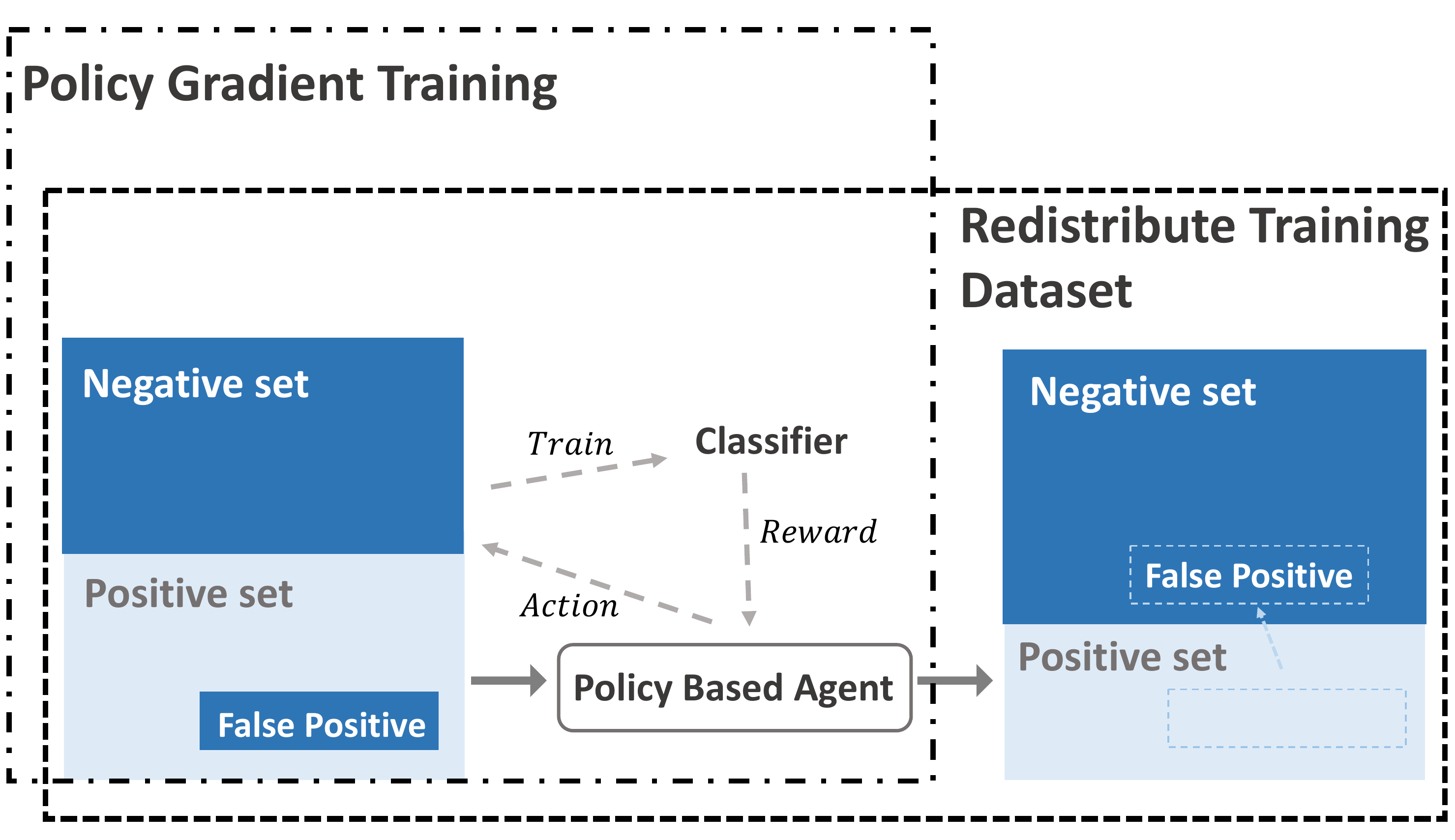}
\caption{Our deep reinforcement learning framework aims at dynamically recognizing false positive samples, and moving them from the positive set to the negative set during distant supervision.}

\label{fig:framework}
\end{center}
\vspace{-3ex}
\end{figure}

A major issue encountered in the early development of relation extraction algorithms is the data sparsity issue---It is extremely expensive, and almost impossible for human annotators to go through a large corpus of millions of sentences to provide a large amount of labeled training instances. Therefore, distant supervision relation extraction~\cite{mintz2009distant,hoffmann2011knowledge,surdeanu2012multi} becomes popular, because it uses entity pairs from knowledge bases to select a set of noisy instances from unlabeled data. In recent years, neural network approaches~\cite{zeng2014relation,zeng2015distant} have been proposed to train the relation extractor under these noisy conditions. To suppress the noisy\cite{roth2013survey}, recent studies~\cite{lin2016neural} have proposed the use of attention mechanisms to place soft weights on a set of noisy sentences, and select samples. However, we argue that only selecting one example or based on soft attention weights are not the optimal strategy: To improve the robustness, we need a systematic solution to make use of more instances, while removing false positives and placing them in the right place.

In this paper, we investigate the possibility of using dynamic selection strategies for robust distant supervision.
More specifically, we design a deep reinforcement learning agent, whose goal is to learn to choose whether to remove or remain the distantly supervised candidate instance based on the performance change of the relation classifier. Intuitively, our agent would like to remove false positives, and reconstruct a cleaned set of distantly supervised instances to maximize the reward based on the classification accuracy. Our proposed method is classifier-independent, and it can be applied to any existing distant supervision model. Empirically, we show that our method has brought consistent performance gains in various deep neural network based models, achieving strong performances on the widely used New York Times dataset~\cite{riedel2010modeling}.
Our contributions are three-fold:
\begin{itemize}
\item We propose a novel deep reinforcement learning framework for robust distant supervision relation extraction.
\item Our method is model-independent, meaning that it could be applied to any state-of-the-art relation extractors.
\item We show that our method can boost the performances of recently proposed neural relation extractors.
\end{itemize}

In Section~\ref{sec:related}, we will discuss related works on distant supervision relation extraction. Next, we will describe our robust distant supervision framework in Section~\ref{sec:method}. In Section~\ref{sec:exp}, empirical evaluation results are shown. And finally, we conclude in Section~\ref{sec:conclude}.

\section{Related Work}
\label{sec:related}
\newcite{mintz2009distant} is the first study that combines dependency path and feature aggregation for distant supervision. However, this approach would introduce a lot of false positives, as the same entity pair might have multiple relations.
To alleviate this issue, \newcite{hoffmann2011knowledge} address this issue, and propose a model to jointly learn with multiple relations. \newcite{surdeanu2012multi} further propose a multi-instance multi-label learning framework to improve the performance. Note that these early approaches do not explicitly remove noisy instances, but rather hope that the model would be able to suppress the noise.

Recently, with the advance of neural network techniques, deep learning methods~\cite{zeng2014relation,zeng2015distant} are introduced, and the hope is to model noisy distant supervision process in the hidden layers. However, their approach only selects one most plausible instance per entity pair, inevitably missing out a lot of valuable training instances. Recently, \newcite{lin2016neural} propose an attention mechanism to select plausible instances from a set of noisy instances. However, we believe that soft attention weight assignment might not be the optimal solution, since the false positives should be completely removed and placed in the negative set. \newcite{ji2017distant} combine the external knowledge to rich the representation of entity pair, in which way to improve the accuracy of attention weights. Even though these above-mentioned methods can select high-quality instances, they ignore the false positive case: all the sentences of one entity pair belongs to the false positives. In this work, we take a radical approach to solve this problem---We will make use of the distantly labeled resources as much as possible, while learning a independent false-positive indicator to remove false positives, and place them in the right place. 
After our ACL submission, we notice that a contemporaneous study \newcite{feng2018reinforcement} also adopts reinforcement learning to learn an instance selector, but their reward is calculated from the prediction probabilities. In contrast, while in our method, the reward is intuitively reflected by the performance change of the relation classifier. 
Our approach is also complement to most of the approaches above, and can be directly applied on top of any existing relation extraction classifiers.

\section{Reinforcement Learning for Distant Supervision}
\label{sec:method}
We introduce a performance-driven, policy-based reinforcement learning method to heuristically recognize false positive samples. Comparing to a prior study that has underutilized the distantly-supervised samples~\cite{lin2016neural}, we consider an RL agent for robust distant supervision relation extraction. 
We first describe the definitions of our RL method, including the policy-based agent, external environment, and pre-training strategy. Next, we describe the retraining strategy for our RL agent. The goal of our agent is to determine whether to {\bfseries retain} or {\bfseries remove} a distantly-supervised sentence, based on the performance change of relation classifier. Finally, we describe the noisy-suppression method, where we teach our policy-based agent to make a redistribution for a cleaner distant supervision training dataset.

Distant supervision relation extraction is to predict the relation type of entity pair under the automatically-generated training set. However, the issue is that these distantly-supervised sentences that mention this entity pair may not express the desired relation type. Therefore, what our RL agent should do is to determine whether the distantly-supervised sentence is a true positive instance for this relation type. 
For reinforcement learning, external environment and RL agent are two necessary components, and a robust agent is trained from the dynamic interaction between these two parts~\cite{arulkumaran2017brief}. 
First, the prerequisite of reinforcement learning is that the external environment should be modeled as a Markov decision process (MDP). However, the traditional setting of relation extraction cannot satisfy this condition: the input sentences are independent of each other. In other words, we cannot merely use the information of the sentence being processed as the state. Thus, we add the information from the early states into the representation of the current state, in which way to model our task as a MDP problem~\cite{fang2017learning}. The other component, RL agent, is parameterized with a policy network $\pi_\theta(s,a)=p(a{\mid}s; \theta)$. The probability distribution of actions $\mathcal{A} = \{a_{remove}, a_{remain}\}$ is calculated by policy network based on state vectors. What needs to be noted is that, Deep Q Network (DQN)~\cite{mnih2013playing} is also a widely-used RL method; however, it is not suitable for our case, even if our action space is small. First, we cannot compute the immediate reward for every operation; In contrast, the accurate reward can only be obtained after finishing processing the whole training dataset. Second, the stochastic policy of the policy network is capable of preventing the agent from getting stuck in an intermediate state. The following subsections detailedly introduce the definitions of the fundamental components in the proposed RL method.
\begin{figure*}[t]
\begin{center}
\includegraphics[width=14cm]{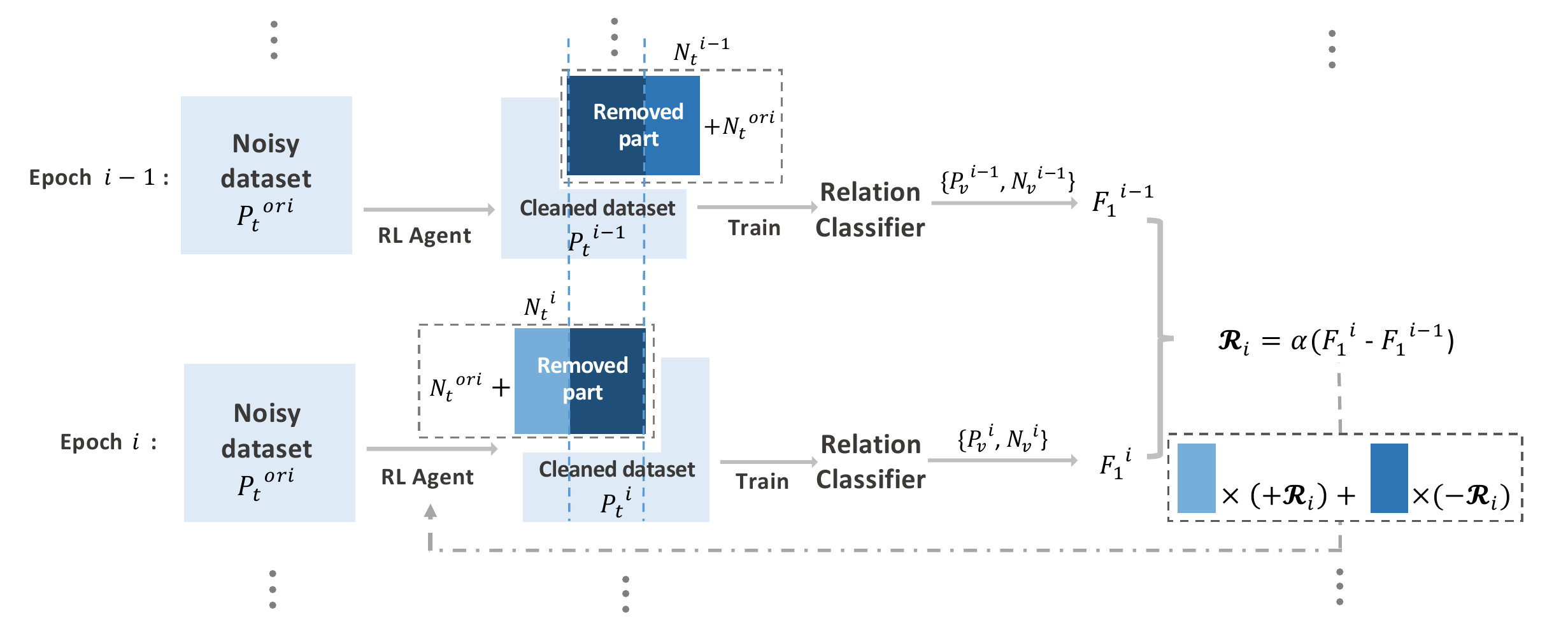}
\caption{\label{framework}The proposed policy-based reinforcement learning framework. The agent tries to remove the wrong-labeled sentences from the distantly-supervised positive dataset $P^{ori}$. In order to calculate the reward, $P^{ori}$ is split into the training part $P_t^{ori}$ and the validation part $P_v^{ori}$; their corresponding negative part are represented as $N_t^{ori}$ and $N_v^{ori}$. In each epoch $i$, the agent performs a series of actions to recognize the false positive samples from $P_t^{ori}$ and treat them as negative samples. Then, a new relation classifier is trained under the new dataset $\{P_t^i,N_t^i\}$. With this relation classifier, $F_1$ score is calculated from the new validation set $\{P_v^i,N_v^i\}$, where $P_v^i$ is also filtered by the current agent. After that, the current reward is measured as the difference of $F_1$ between the adjacent epochs.}
\end{center}
\end{figure*}

\paragraph{States}
In order to satisfy the condition of MDP, the state $s$ includes the information from the current sentence and the sentences that have been removed in early states. The semantic and syntactic information of sentence is represented by a continuous real-valued vector. According to some state-of-the-art supervised relation extraction approaches~\cite{zeng2014relation,nguyen2015event}, we utilize both word embedding and position embedding to convert sentence into vector.
With this sentence vector, the current state is the concatenation of the current sentence vector and the average vector of the removed sentences in early states. We give relatively larger weight for the vector of the current sentence, in which way to magnify the dominating influence of the current sentence information for the decision of action.

\paragraph{Actions}
At each step, our agent is required to determine whether the instance is false positive for target relation type. Each relation type has a agent\footnote{We also tried the strategy that just builds a single agent for all relation types: a binary classifier(TP/FP) or a multi-class classifier(rela1/rela2/.../FP). But, it has the limitation in the performance. We found that our one-agent-for-one-relation strategy obtained better performance than the single agent strategy.}. There are two actions for each agent: whether to remove or retain the current instance from the training set. With the initial distantly-supervised dataset that is blended with incorrectly-labeled instances, we hope that our agent is capable of using the policy network to filter noisy instances; Under this cleaned dataset, distant supervision is then expected to achieve better performance. 

\paragraph{Rewards}
As previously mentioned, the intuition of our model is that, when the incorrectly-labeled instances are filtered, the better performance of relation classifier will achieve. Therefore, we use the change of performance as the result-driven reward for a series of actions decided by the agent. Compared to accuracy, we adopt the $F_1$ score as the evaluation criterion, since accuracy might not be an indicative metric in a multi-class classification setting where the data distribution could be imbalanced. Thus, the reward can be formulated as the difference between the adjacent epochs:
\begin{equation}
\mathnormal{R_i} = \alpha(F^i_1-F^{i-1}_1)
\label {reward}
\end{equation}
As this equation shows, in step $i$, our agent is given a positive reward only if $F_1$ gets improved; otherwise, the agent will receive a negative reward. Under this setting, the value of reward is proportional to the difference of $F_1$, and $\alpha$ is used to convert this difference into a rational numeric range. Naturally, the value of the reward is in a continuous space, which is more reasonable than a binary reward ($-1$ and $1$), because this setting can reflect the number of wrong-labeled instance that the agent has removed. In order to avoid the randomness of $F_1$, we use the average $F_1$ of last five epochs to calculate the reward.

\paragraph{Policy Network}
For each input sentence, our policy network is to determine whether it expresses the target relation type and then make removal action if it is irrelevant to the target relation type. Thus, it is analogous to a binary relation classifier. CNN is commonly used to construct relation classification system~\cite{santos2015classifying, xu2015semantic,shen-huang:2016:COLING}, so we adopt a simple CNN with window size $c_w$ and kernel size $c_k$, to model policy network $\pi(s;\theta)$. The reason why we do not choice the variants of CNN~\cite{zeng2015distant,lin2016neural} that are well-designed for distant supervision is that these two models belong to bag-level models (dealing with a bag of sentences simultaneously) and deal with the multi-classification problem; We just need a model to do binary sentence-level classification. Naturally, the simpler network is adopted.

\subsection{Training Policy-based Agent}
\label{train_agent}
Unlike the goal of distant supervision relation extraction, our agent is to determine whether an annotated sentence expresses the target relation type rather than predict the relationship of entity pair, so sentences are treated independently despite belonging to the same entity pair. In distant supervision training dataset, one relation type contains several thousands or ten thousands sentences; moreover, reward $\mathnormal{R}$ can only be calculated after processing the whole positive set of this relation type. If we randomly initialize the parameters of policy network and train this network by trial and errors, it will waste a lot of time and be inclined to poor convergence properties. In order to overcome this problem, we adopt a supervised learning procedure to pre-train our policy network, in which way to provide a general learning direction for our policy-based agent.

\subsubsection{Pre-training Strategy}
The pre-training strategy, inspired from $\mathnormal{AlphaGo}$~\cite{silver2016mastering}, is a common strategy in RL related works to accelerate the training of RL agents. Normally, they utilize a small part of the annotated dataset to train policy networks before reinforcement learning. For example, $\mathnormal{AlphaGo}$ uses the collected experts moves to do a supervised learning for $\mathnormal{Go}$ RL agent. However, in distant supervision relation extraction task, there is not any supervised information that can be used unless let linguistic experts to do some manual annotations for part of the entity pairs. However, this is expensive, and it is not the original intention of distant supervision. Under this circumstance, we propose a compromised solution. 
With well-aligned corpus, the true positive samples should have evident advantage in quantity compared with false positive samples in the distantly-supervised dataset. 
So, for a specific relation type, we directly treat the distantly-supervised positive set as the positive set, and randomly extract part of distantly-supervised negative set as the negative set.
In order to better consider prior information during this pre-training procedure, the amount of negative samples is 10 times of the number of positive samples. It is because, when learning with massive negative samples, the agent is more likely to develop toward a better direction. Cross-entropy cost function is used to train this binary classifier, where the negative label corresponds to the removing action, and the positive label corresponds to the retaining action.
\begin{dmath}
\mathnormal{J}(\theta) = \sum_{i} {y_i}{\log[\pi(a=y_i{\mid}s_i;\theta)]}+(1-{y_i}){\log[1-\pi(a=y_i{\mid}s_i;\theta)]}
\label{pretrain}
\end{dmath}
Due to the noisy nature of the distantly-labeled instances, if we let this pre-training process overfit this noisy dataset, the predicted probabilities of most samples tend to be close to 0 or 1, which is difficult to be corrected and unnecessarily increases the training cost of reinforcement learning. So, we stop this training process when the accuracy reaches $85\%\sim90\%$. Theoretically, our approach can be explained as increasing the entropy of the policy gradient agent, and preventing the entropy of the policy being too low, which means that the lack of exploration may be a concern.

    \begin{algorithm*}[t]  
        \caption{Retraining agent with rewards for relation $k$. For a clearer expression, $k$ is omitted in the following algorithm.}  
        \begin{algorithmic}[1]  
            \Require Positive set $\{\mathnormal{P_t^{ori}},\mathnormal{P_v^{ori}}\}$, Negative set $\{\mathnormal{N_t^{ori}},\mathnormal{N_v^{ori}}\}$, the fixed number of removal $\gamma_t,\gamma_v$ 
            \State Load parameters $\theta$ from pre-trained policy network
            \State Initialize $s^*$ as the all-zero vector with the same dimension of $s_j$
            \For{epoch $i = 1 \to N$}
                \For{$s_j \in \mathnormal{P_t^{ori}}$}
                	\State $\widetilde{s_j}=concatenation(s_j,s^*)$
                	\State Randomly sample $a_j\sim\pi(a{\mid}\widetilde{s_j};\theta)$; compute $p_j=\pi(a=0{\mid}\widetilde{s_j};\theta)$
                    \If {$a_j == 0$}
                    	\State Save tuple $t_j=(\widetilde{s_j}, p_j)$ in $\mathnormal{T}$ and recompute the average vector of removed sentences $s^*$
                    \EndIf
                \EndFor
                \State Rank $\mathnormal{T}$ based on $p_j$ from high to low, obtain $\mathnormal{T_{rank}}$
            	\For{$t_i$ in $\mathnormal{T_{rank}}[:\gamma_t]$}
            		\State Add $t_i[0]$ into $\Psi_{i}$
            	\EndFor
                \State $\mathnormal{P_t^i} = \mathnormal{P_t^{ori}}-\Psi_{i}, \mathnormal{N_t^i} = \mathnormal{N_t^{ori}}+\Psi_{i}$, and generate the new validation set $\{\mathnormal{P_v^i}, \mathnormal{N_v^i}\}$ with current agent
                \State Train the relation classifier based on $\{\mathnormal{P_t^i},\mathnormal{N_t^i}\}$
            	\State Calculate $F_1^i$ on the new validation set $\{\mathnormal{P_v^i},\mathnormal{N_v^i}\}$, and Save $F_1^i$, $\Psi_{i}$
                \State $\mathcal{R}=\alpha(F_1^i-F_1^{i-1})$
                \State $\Omega_{i-1} = \Psi_{i-1} - \Psi_{i}\cap\Psi_{i-1}; \, \Omega_{i} = \Psi_{i} - \Psi_{i}\cap\Psi_{i-1}$
            	\State 
             \State Updata $\theta$: $g\propto{\bigtriangledown_{\theta}}\sum^{\Omega_{i}}{\log\pi(a{\mid}s;\theta)}\mathcal{R} + {\bigtriangledown_{\theta}}\sum^{\Omega_{i-1}}{\log\pi(a{\mid}s;\theta)}(\mathcal{-R})$
            \EndFor
        \end{algorithmic}  
    \end{algorithm*}

\subsubsection{Retraining Agent with Rewards}
As shown in Figure~\ref{framework}, in order to discover incorrectly-labeled instances without any supervised information, we introduce a policy-based RL method. What our agent tries to deal with is the noisy samples from the distantly-supervised positive dataset; Here we call it as the \emph{DS positive dataset}. We split it into the training positive set $P_t^{ori}$ and the validation positive set $P_v^{ori}$; naturally, both of these two set are noisy. Correspondingly, the training negative set $N_t^{ori}$ and the validation negative set $N_v^{ori}$ are constructed by randomly selected from the DS negative dataset.
In every epoch, the agent removes a noisy sample set $\Psi_{i}$ from $P_t^{ori}$ according to the stochastic policy $\pi(a{\mid}s)$, and we obtain a new positive set $P_t = P_t^{ori}-\Psi_{i}$. Because $\Psi_{i}$ is recognized as the wrong-labeled samples, we redistribute it into the negative set $N_t = N_t^{ori}+\Psi_{i}$. Under this setting, the scale of training set is constant for each epoch.
Now we utilize the cleaned data $\{P_t,N_t\}$ to train a relation classifier. The desirable situation is that RL agent has the capacity to increase the performance of relation classifier through relocating incorrectly-labeled false positive instances. Therefore, we use the validation set $\{P_v^{ori},N_v^{ori}\}$ to measure the performance of the current agent. First, this validation set is filtered and redistributed by the current agent as $\{P_v,N_v\}$; the $F_1$ score of the current relation classifier is calculated from it.
Finally, the difference of $F_1$ scores between the current and previous epoch is used to calculate reward. Next, we will introduce several strategies to train a more robust RL agent.


\paragraph{Removing the fixed number of sentences in each epoch}
In every epoch, we let the RL agent to remove a fixed number of sentences or less (when the number of the removed sentences in one epoch does not reach this fixed number during training), in which way to prevent the case that the agent tries to remove more false positive instances by removing more instances. Under the restriction of fixed number, if the agent decides to remove the current state, it means the chance of removing other states decrease. Therefore, in order to obtain a better reward, the agent should try to remove a instance set that includes more negative instances.

\paragraph{Loss function}
The quality of the RL agent is reflected by the quality of the removed part. After the pre-training process, the agent just possesses the ability to distinguish the obvious false positive instances, which means the discrimination of the indistinguishable wrong-labeled instances are still ambiguous. Particularly, this indistinguishable part is the criterion to reflect the quality of the agent. Therefore, regardless of these easy-distinguished instances, the different parts of the removed parts in different epochs are the determinant of the change of $F_1$ scores. Therefore, we definite two sets:
\begin{align}
\Omega_{i-1} &= \Psi_{i-1} - (\Psi_{i}\cap\Psi_{i-1})\\
  \Omega_{i} &= \Psi_{i} - (\Psi_{i}\cap\Psi_{i-1})
\label {different_part}
\end{align}
where $\Psi_{i}$ is the removed part of epoch $i$. $\Omega_{i-1}$ and $\Omega_{i}$ are represented with the different colors in Figure~\ref{framework}. If $F_1$ score increases in the epoch $i$, it means the actions of the epoch $i$ is more reasonable than that in the epoch $i-1$. In other words, $\Omega_{i}$ is more negative than $\Omega_{i-1}$. Thus, we assign the positive reward to $\Omega_{i}$ and the negative reward to $\Omega_{i-1}$, and vice versa. In summary, the ultimate loss function is formulated as follow:
\begin{dmath}
\mathnormal{J}(\theta) = \sum^{\Omega_{i}}{\log\pi(a{\mid}s;\theta)}\mathnormal{R} + \sum^{\Omega_{i-1}}{\log\pi(a{\mid}s;\theta)}(\mathnormal{-R})
\label{RL_loss}
\end{dmath}

\subsection{Redistributing Training Dataset with Policy-based Agents}
Through the above reinforcement learning procedure, for each relation type, we obtain a agent as the false-positive indicator. These agents possess the capability of recognizing incorrectly-labeled instances of the corresponding relation types. We adopt these agents as classifiers to recognize false positive samples in the noisy distantly-supervised training dataset. For one entity pair, if all the sentence aligned from corpus are classified as false positive, then this entity pair is redistributed into the negative set.

\section{Experiments}
\label{sec:exp}
We adopt a policy-based RL method to generate a series of relation indicators and use them to redistribute training dataset by moving false positive samples to negative sample set. Therefore, our experiments are intended to demonstrate that our RL agents possess this capability. 

\subsection{Datast and Evaluation Metrics}
We evaluate the proposed method on a commonly-used dataset\footnote{http://iesl.cs.umass.edu/riedel/ecml/}, which is first presented in~\newcite{riedel2010modeling}. This dataset is generated by aligning entity pairs from Freebase with New York Times corpus(NYT). Entity mentions of NYT corpus are recognized by the Stanford named entity recognizer~\cite{finkel2005incorporating}. The sentences from the years 2005-2006 are used as the training corpus and sentences from 2007 are used as the testing corpus. 
There are 52 actual relations and a special relation $NA$ which indicates there is no relation between the head and tail entities. 
The sentences of $NA$ are from the entity pairs that exist in the same sentence of the actual relations but do not appear in the Freebase.

Similar to the previous works, we adopt the held-out evaluation to evaluate our model, which can provide an approximate measure of the classification ability without costly human evaluation. Similar to the generation of the training set, the entity pairs in test set are also selected from Freebase, which will be predicted under the sentences discovered from the NYT corpus.

\subsection{Experimental Settings}

\subsubsection{Policy-based Agent}
The action space of our RL agent just includes two actions. Therefore, the agent can be modeled as a binary classifier. We adopt a single-window CNN as this policy network. The detailed hyperparameter settings are presented in Table~\ref{paraset}. As for word embeddings, we directly use the word embedding file released by~\newcite{lin2016neural}\footnote{https://github.com/thunlp/NRE}, which just keeps the words that appear more than 100 times in NYT. Moreover, we have the same dimension setting of the position embedding, and the maximum length of relative distance is $-30$ and $30$ (``-'' and ``+'' represent the left and right side of the entities). The learning rate of reinforcement learning is $2e^{-5}$. For each relation type, the fixed number $\gamma_t,\gamma_v$ are according to the pre-trained agent. When one relation type has too many distant-supervised positive sentences (for example, ‘/location/location/contains’ has 75768 sentences), we sample a subset of size 7,500 sentences to train the agent. For the average vector of the removed sentences, in the pre-training process and the first state of the retraining process, it is set as all-zero vector.

\subsubsection{Relation Classifier for Calculating Reward}
In order to evaluate a series of actions by agent, we use a simple CNN model, because the simple network is more sensitive to the quality of the training set. The proportion between $P_t^{ori}$ and $P_v^{ori}$ is 2:1, and they are all derived from the training set of Riedel dataset; the corresponding negative sample sets $N_t^{ori}$ and $N_v^{ori}$ are randomly selected from the Riedel negative dataset, whose size is twice that of their corresponding positive sets. 

\begin{table}[t]
\centering
\begin{tabular}{cc}
\hline
\hline 
\bf {Hyperparameter} & \bf {Value} \\
\hline
Window size $c_w$ & 3  \\
Kernel size $c_k$ & 100\\
Batch size & 64  \\
Regulator $\alpha$ & 100 \\
\hline
\hline
\end{tabular}
\caption{\label{paraset} Hyperparameter settings.}
\end{table}

\begin{table}[t]
\small
\centering 
\begin{tabular}{lcccc}
\hline
\hline 
\bf {\bf ID} & {\bf Relation}& {\bf Original} &{\bf Pretrain} & {\bf RL}\\
\hline
1 & /peo/per/pob & 55.60 &53.63 & {\bf55.74} \\
2 & /peo/per/n & 78.85 & 80.80 & {\bf83.63} \\
3 & /peo/per/pl & 86.65 &89.62 & {\bf90.76} \\
4 & /loc/loc/c & 80.78 & 83.79 & {\bf85.39} \\
5 & /loc/cou/ad & {\bf90.9} & 88.1 & 89.86 \\
6 & /bus/per/c & 81.03 &82.56 & {\bf84.22} \\
7 & /loc/cou/c & 88.10 & 93.78 & {\bf 95.19} \\
8 & /loc/adm/c & 86.51 & 85.56 & {\bf86.63} \\
9 & /loc/nei/n & 96.51 & 97.20 & {\bf 98.23} \\
10 & /peo/dec/p & 82.2 & 83.0 & {\bf84.6} \\
\hline
\hline
\end{tabular}
\caption{\label{mid_result} Comparison of $F_1$ scores among three cases: the relation classifier is trained with the original dataset, the redistributed dataset generated by the pre-trained agent, and the redistributed dataset generated by our RL agent respectively. The name of relation types are abbreviated: $/peo/per/pob$ represents $/people/person/place\_of\_birth$}
\end{table}

\subsection{The Effectiveness of Reinforcement Learning}
In Table~\ref{mid_result}, we list the $F_1$ scores before and after adopting the proposed RL method. Even though there are 52 actual relation types in Riedel dataset, only 10 relation types have more than 1000 positive instances\footnote{The supervised relation classification task Semeval-2010 Task 8~\cite{hendrickx2009semeval} annotates nearly 1,000 instances for each relation type.}. Because of the randomness of deep neural network on the small-scale dataset, we just train policy-based agents for these 10 relation types. First, compared with \emph Original case, most of the \emph{Pretrain} agents yield obvious improvements: It not only demonstrates the rationality of our pre-training strategy, but also verifies our hypothesis that most of the positive samples in Riedel dataset are true positive. More significantly, after retraining with the proposed policy-based RL method, the $F_1$ scores achieve further improvement, even for the case the \emph{Pretrain} agents perform bad. These comparable results illustrate that the proposed policy-based RL method is capable of making agents develop towards a good direction.

\begin{figure}[t]
\small
\begin{center}
\includegraphics[width=6cm]{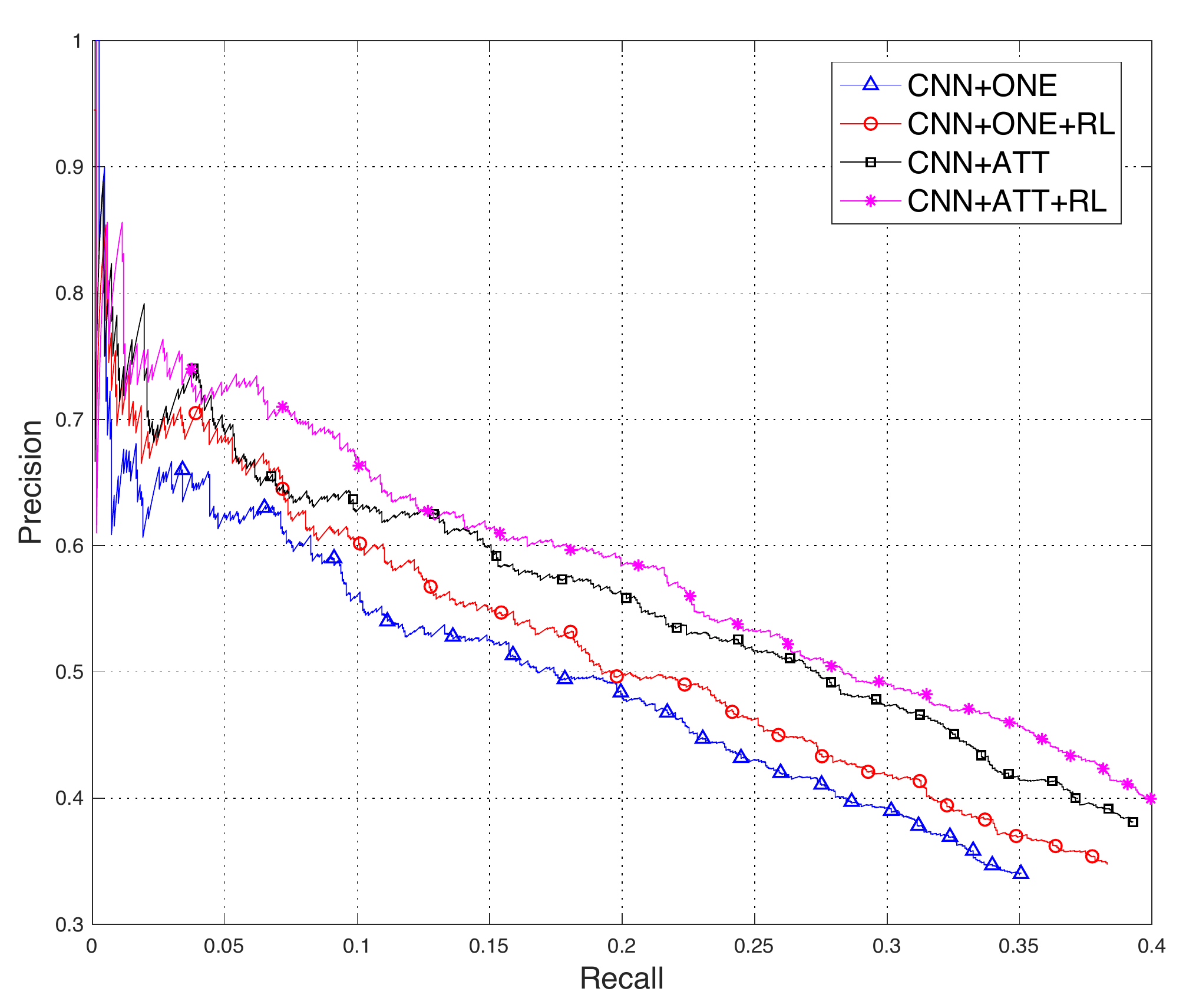}
\caption{\label{CNN_curve}Aggregate PR curves of CNN_based model.}
\end{center}
\end{figure}

\begin{figure}[t]
\begin{center}
\includegraphics[width=6cm]{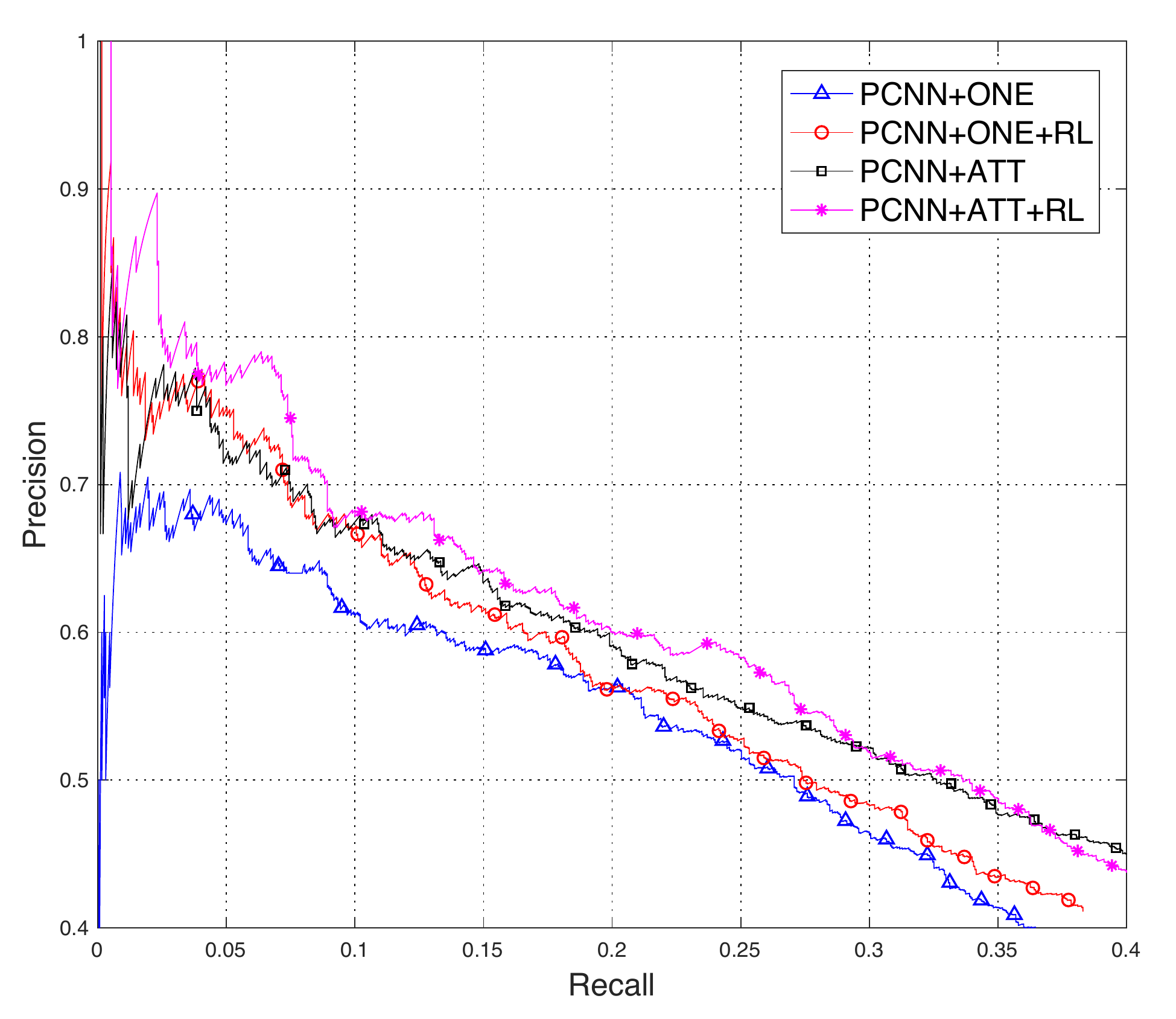}
\caption{\label{PCNN_curve}Aggregate PR curves of PCNN_based model.}
\end{center}
\end{figure}

\subsection{Impact of False Positive Samples}
~\newcite{zeng2015distant} and~\newcite{lin2016neural} are both the robust models to solve wrong labeling problem of distant supervision relation extraction.~\newcite{zeng2015distant} combine at-least-one multi-instance learning with deep neural network to extract only one active sentence to predict the relation between entity pair;~\newcite{lin2016neural} combine all sentences of one entity pair and assign soft attention weights to them, in which way to generate a compositive relation representation for this entity pair. However, the false positive phenomenon also includes the case that all the sentences of one entity pair are wrong, which is because the corpus is not completely aligned with the knowledge base. This phenomenon is also common between Riedel dataset and Freebase through our manual inspection. Obviously, there is nothing the above two methods can do in this case. 

The proposed RL method is to tackle this problem. We adopt our RL agents to redistribute Riedel dataset by moving false positive samples into the negative sample set. Then we use~\newcite{zeng2015distant} and~\newcite{lin2016neural} to predict relations on this cleaned dataset, and compare the performance with that on the original Riedel dataset. As shown in Figure~\ref{CNN_curve} and Figure~\ref{PCNN_curve}, under the assistant of our RL agent, the same model can achieve obvious improvement with more reasonable training dataset. In order to give the more intuitive comparison, we calculate the AUC value of each PR curve, which reflects the area size under these curves. These comparable results also indicate the effectiveness of our policy-based RL method. Moreover, as can be seen from the result of t-test evaluation, all the p-values are less than 5e-02, so the improvements are significant.

\begin{table}[t]
\normalsize
\centering
\begin{tabular}{lccc}
\hline
\hline 
\bf {Model} & \bf {-} & \bf {+RL} & \bf {p-value} \\
\hline
CNN+ONE & 0.177 & \bf {0.190} & 1.24e-4 \\
CNN+ATT & 0.219 & \bf {0.229} & 7.63e-4 \\
PCNN+ONE & 0.206 &\bf {0.220} & 8.35e-6 \\
PCNN+ATT & 0.253 & \bf {0.261} & 4.36e-3\\
\hline
\hline
\end{tabular}
\caption{\label{AUC} Comparison of AUC values between previous studies and our RL method, and the p-value of t-test.}
\end{table}

\begin{table*}[t]
\centering
\begin{tabular}{p{2cm}p{13cm}}
\hline
\hline 
\bf {Relation} & \bf {/people/person/place\_of\_birth} \\
\hline
FP & 1. GHETTO SUPERSTAR ( THE MAN THAT I AM) -- Ranging from {\bf Pittsburgh} to Broadway, {\bf Billy Porter} performs his musical memoir.\\
\hline
FP & 1. ``They are trying to create a united front at home in the face of the pressures {\bf Syria} is facing,`` said {\bf Sami Moubayed}, a political analyst and writer here.\\
 & 2. ``Iran injected {\bf Syria} with a lot of confidence: stand up, show defiance,`` said {\bf Sami Moubayed}, a political analyst and writer in Damascus. \\
\hline
\hline
\bf {Relation} & \bf {/people/deceased\_person/place\_of\_death} \\
\hline
FP & 1. Some {\bf New York city} mayors -- {\bf William O'Dwyer}, Vincent R. Impellitteri and Abraham Beame -- were born abroad.\\
 & 2. Plenty of local officials have, too, including two {\bf New York city} mayors, James J. Walker, in 1932, and {\bf William O'Dwyer}, in 1950.\\
\hline
\hline
\end{tabular}
\caption{\label{case_study} Some examples of the false positive samples detected by our policy-based agent. Each row denotes the annotated sentences of one entity pair.}
\end{table*}
\subsection{Case Study}
\begin{figure}[t]
\begin{center}
\includegraphics[width=7cm]{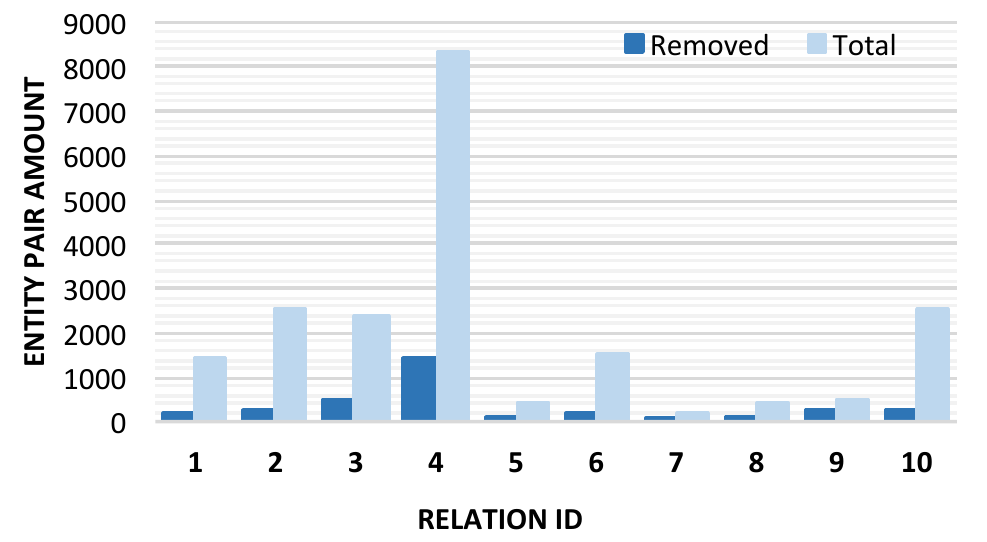}
\caption{\label{remove_sent} This figure presents the scale of the removed part for each relation type, where the horizontal axis corresponds to the IDs in Table~\ref{mid_result}.}
\end{center}
\end{figure}

Figure~\ref{remove_sent} indicates that, for different relations, the scale of the detected false positive samples is not proportional to the original scale, which is in accordance with the actual accident situation. At the same time, we analyze the correlation between the false positive phenomenon and the number of sentences of entity pairs : With this the number ranging from 1 to 5, the corresponding percentages are [55.9\%, 32.0\%, 3.7\%, 4.4\%, 0.7\%]. This distribution is consistent with our assumption. Because Freebase is, to some extent, not completely aligned with the NYT corpus, entity pairs with fewer sentences are more likely to be false positive, which is the major factor hindering the performance of the previous systems. 
In Table~\ref{case_study}, we present some false positive examples selected by our agents. Taking entity pair (\emph{Sami Moubayed}, \emph{Syria}) as an example, it is obvious that there is not any valuable information reflecting relation \emph{/people/person/place\_of\_birth}. Both of these sentences talks about the situation analysis of \emph{Syria} from the political analyst \emph{Sami Moubayed}. We also found that, for some entity pairs, even though there are multiple sentences, all of them are identical. This phenomenon also increases the probability of the appearance of false positive samples.

\section{Conclusion}
\label{sec:conclude}
In this work, we propose a deep reinforcement learning framework for robust distant supervision. 
The intuition is that, in contrast to prior works that utilize only one instance per entity pair and use soft attention weights to select plausible distantly supervised examples, we describe a policy-based framework to systematically learn to relocate the false positive samples, and better utilize the unlabeled data. More specifically, our goal is to teach the reinforcement agent to optimize the selection/redistribution strategy that maximizes the reward of boosting the performance of relation classification. An important aspect of our work is that our framework does not depend on a specific form of the relation classifier, meaning that it is a plug-and-play technique that could be potentially applied to any relation extraction pipeline. In experiments, we show that our framework boosts the performance of distant supervision relation extraction of various strong deep learning baselines on the widely used New York Times - Freebase dataset.

\section*{Acknowledge}
This work was supported by National Natural Science Foundation of China (61702047), Beijing Natural Science Foundation (4174098), the Fundamental Research Funds for the Central Universities (2017RC02) and National Natural Science Foundation of China (61703234)

\bibliography{acl2018}
\bibliographystyle{acl_natbib}

\end{document}